\documentclass[conference]{IEEEtran}
\IEEEoverridecommandlockouts
\usepackage{cite}
\usepackage{amsmath,amssymb,amsfonts}
\usepackage{algorithmic}
\usepackage{graphicx}
\usepackage{amsmath,amsfonts}
\usepackage{tikz}
\usepackage{pgfplots}
\usetikzlibrary{shapes, arrows.meta, positioning, calc, backgrounds, patterns, fadings, shadows}
\usepackage{multirow}
\usepackage{algorithm}
\usepackage{textcomp}
\usepackage{booktabs}
\usepackage{xcolor}
\usepackage[table]{xcolor}
\definecolor{lightred}{rgb}{1, 0.76, 0.76}
\definecolor{lightgreen}{rgb}{0.88, 1, 0.88}
\definecolor{lightorange}{rgb}{1, 0.93, 0.76}

\definecolor{lightyellow}{rgb}{0.88, 0.95, 1.0}
\usepackage[textsize=tiny]{todonotes}
\def\BibTeX{{\rm B\kern-.05em{\sc i\kern-.025em b}\kern-.08em
    T\kern-.1667em\lower.7ex\hbox{E}\kern-.125emX}}
\begin{document}

\title{GP-GS: Gaussian Processes Densification for 3D Gaussian Splatting}

\author{
Zhihao Guo$^{1,*}$ \quad
Jingxuan Su$^{2,*}$ \quad
Chenghao Qian$^{3}$ \quad
Shenglin Wang $^{4}$ \quad
Jinlong Fan $^{5}$ \quad
Jing Zhang $^{6}$ \quad 
Wei Zhou $^{7}$ \quad \\
Hadi Amirpour $^{8}$ \quad
Yunlong Zhao $^{9}$ \quad
Liangxiu Han $^{1}$ \quad
Peng Wang$^{1,\dag}$ \\
$^1$ Manchester Metropolitan University \\
$^2$ SECE, Peking University \\
$^3$ University of Leeds \\
$^4$ Pengcheng Laboratory \\
$^5$ Hangzhou Dianzi University \\
$^6$ Wuhan University \\
$^7$ Cardiff University \\
$^8$ University of Klagenfurt \\
$^9$ Imperial College London
}
\maketitle

\begin{abstract} 3D Gaussian Splatting (3DGS) enables photorealistic rendering but suffers from artefacts due to sparse Structure-from-Motion (SfM) initialisation. To address this limitation, we propose GP-GS, a Gaussian Process (GP) based densification framework for 3DGS optimisation.
GP-GS formulates point cloud densification as a continuous regression problem, where a GP learns a local mapping from 2D pixel coordinates to 3D position and colour attributes. An adaptive neighbourhood-based sampling strategy generates candidate pixels for inference, while GP-predicted uncertainty is used to filter unreliable predictions, reducing noise and preserving geometric structure. Extensive experiments on synthetic and real-world benchmarks demonstrate that GP-GS consistently improves reconstruction quality and rendering fidelity, achieving up to 1.12 dB PSNR improvement over strong baselines.\end{abstract}


\section{Introduction}
\label{Introduction}

Novel View Synthesis (NVS) is a fundamental yet challenging problem in computer vision and computer graphics, aiming to generate novel viewpoints of a given scene from multi-view observations. It plays a critical role in applications such as digital twinning~\cite{wang2024deep,qian2025weatheredit}, virtual reality~\cite{yan20243dsceneeditor,qian2025weathergs}, and robotics~\cite{wang2024robot,yan2024renderworld}. Neural Radiance Fields (NeRF)~\cite{mildenhall2021nerf} have revolutionised NVS by implicitly modelling volumetric scene representations, achieving high-fidelity rendering without explicit 3D geometry reconstruction. However, NeRF-based methods suffer from computational inefficiency and the requirement of dense sampling along rays, leading to slow inference speeds despite recent acceleration efforts~\cite{roessle2022dense}.

3D Gaussian Splatting (3DGS)~\cite{kerbl20233d} and its variants~\cite{yang2024gaussian,bulo2024revising,lu2024scaffold} have emerged as promising alternatives for real-time rendering. Unlike NeRF, which rely on ray-marching and volumetric integration, 3DGS represents scenes explicitly using a set of 3D Gaussians with learnable attributes like positions and colors. The rasterisation-based splatting strategy can be considered a forward rendering technique that avoids costly ray sampling and enables efficient parallel rendering, making it a compelling choice for NVS applications. 
However, existing 3DGS pipelines rely on SfM, which often yields sparse point clouds in texture-less or clustered regions. This poor initialisation leads to rendering artefacts and a significant loss of fine detail. To compensate, Adaptive Density Control (ADC) dynamically clone or prune Gaussians, but lacks geometric priors, often yielding noisy distributions that fail to preserve scene structure. This constrains fidelity, highlighting the need for a more structured refinement strategy. While learning-based approaches~\cite{chung2024depth,xu2025depthsplat,cheng2024gaussianpro } have been explored for point cloud densification, they typically require large-scale training data, introduce additional network parameters, and provide limited uncertainty estimates. In contrast, GP offer a data-efficient, non-parametric alternative that naturally models predictive uncertainty and enforces local smoothness. These properties make GP particularly well-suited for interpolating sparse SfM point clouds, where reliable uncertainty estimation and controlled local densification are critical for avoiding over-densification and geometric noise. In this paper, we propose GP-GS, a framework that enhances 3DGS initialisation via uncertainty-guided densification. We formulate this task as a continuous regression problem, where the GP learns a mapping from 2D image pixels to a denser point cloud enriched with 3D position and colour information. To ensure robust densification, we propose an adaptive neighbourhood sampling strategy to select candidate pixels for GP inference. For each candidate, the GP predicts 3D point cloud attributes (position, colour, variance), which are then refined via variance-based filtering. This step eliminates high-uncertainty predictions, reducing noise and providing high-quality initialisation points to enhance 3DGS reconstruction.

Our \textbf{contributions} can be summarised as: 1) We propose a GP model to densify sparse SfM point clouds by learning mappings from 2D image pixels to 3D positions and colours, with uncertainty awareness. 2) We introduce an adaptive neighbourhood-based sampling strategy that generates candidate inputs for GP for 3D points prediction, followed by variance-based filtering to remove high uncertainty predictions, ensuring geometric consistency and enhancing rendering quality. 3) Our GP framework can be seamlessly integrated into existing SfM-based 3DGS pipelines, serving as a flexible plug-and-play module that improves the rendering quality.

\section{Related Work}

\subsection{Gaussian Processes}
\label{gps}
Gaussian Processes (GP) are a collection of random variables, any finite number of which are subject to a joint Gaussian distribution~\cite{rasmussen2003gaussian}. GP are particularly effective in handling sparse data, making them well-suited for scenarios with limited observations~\cite {wang2021computationally}. Their flexibility and built-in uncertainty quantification enable robust predictions and uncertainty-informed signal processing, which has led to their widespread adoption in various computer vision tasks~\cite{zhu2021convolutional}. 

\subsection{3D Gaussian Splatting and Initialisation}
\label{sec:related_work_3dgs}

3DGS~\cite{kerbl20233d} employs a forward rendering approach. By projecting 3D Gaussians onto the 2D image plane via efficient $\alpha$-blending, it achieves real-time rendering speeds with high fidelity. While recent works have improved 3DGS by refining point management strategies~\cite{yang2024gaussian,bulo2024revising,lu2024scaffold}, the fundamental quality of the reconstruction remains heavily contingent on the \textit{initial point cloud density}. A sparse or inaccurate initialisation often leads to local minima, floating artifacts, and blurred geometry, prompting a surge of research into enhanced initialisation strategies.

To overcome the sparsity of SfM point clouds, methods such as Depth-Regularised Optimisation~\cite{chung2024depth} and DepthSplat~\cite{xu2025depthsplat} typically utilise monocular depth maps as a regularisation term in the loss function. While effective, these approaches are computationally inefficient due to the reliance on heavy depth estimation networks, and their performance is inherently limited by the accuracy of the depth priors. Alternatively, rather than relying on external depth estimators, GaussianPro~\cite{cheng2024gaussianpro} adopts a geometric approach, applying a progressive propagation strategy to guide
the densification of 3D Gaussians. A parallel line of research, such as InstantSplat~\cite{fan2024instantsplat}, attempts to bypass the SfM pipeline entirely. These methods leverage powerful pre-trained priors (e.g., MASt3R~\cite{leroy2024grounding}) to initialise both geometry and camera poses directly from images. While promising, they introduce a heavy dependency on the generalisation capability of the pre-trained priors and often require heuristic filtering to manage redundancy. In contrast, our work builds upon the robust, interpretable, and data-driven foundation of the standard SfM pipeline, aiming to resolve its sparsity limitations without discarding its geometric rigour.

\begin{figure*}[h]
    \centering
    \includegraphics[width=1\linewidth]{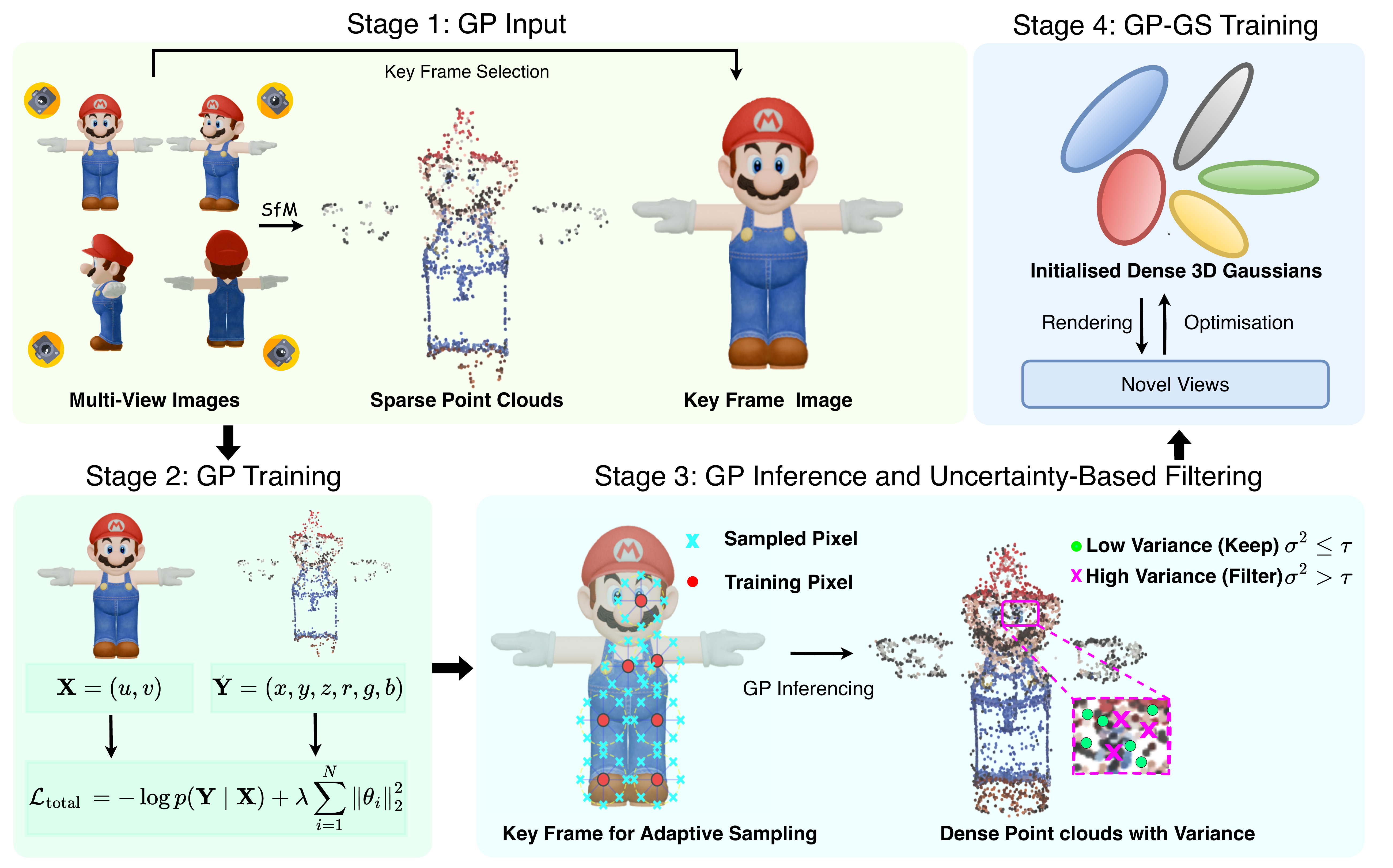}
    \caption{\textbf{The overview of GP-GS.} \textbf{(a)} GP Input: Sparse point cloud is initially obtained using SfM. We then perform pixel-to-point cloud matching and select the key frame that contributes the most to the point cloud. \textbf{(b)} Point Cloud Densification: GP is trained to take key frame pixel coordinates $\mathbf{X}=[u,v]$  and predicts dense point clouds $\mathbf{Y}=[x,y,z,r,g,b]$ with uncertainty (variance). The loss function ensures an optimal mapping between input pixels and point clouds. \textbf{(c)} GP Testing: The adaptive sampling strategy adaptively generate sampling pixels within the image domain, and uncertainty-based filtering removes unreliable predictions. This results in a refined dense point cloud. \textbf{(d)} GP-GS Training: The sparse and the predicted point clouds are merged and then used to initialise dense 3D Gaussians, which are then optimised to refine geometric details. The final rendered novel views demonstrate the effectiveness of GP-GS in reconstructing fine details while maintaining structural coherence.}
    \label{fig:overview}
\end{figure*}

\section{Methodology}


\subsection{Multi-Output Gaussian Process}
\label{Multi Output Gaussian Process}

\textbf{Problem Definition.} We consider a GP regression problem, where the goal is to predict dense point clouds given image pixels. Given a set of $n$ RGB images $\mathcal{I}=\{I_i\}_{i=1}^n$, we can extract the corresponding sparse point clouds \( \mathcal{P}=\{\mathbf{P}_j\}_{j=1}^N \) with a total number of \( N \) points. Each point \( \mathbf{P}_j \) in the point clouds contains its 3D spatial position and the associated colour information: $\mathbf{P}_j = [x_j, y_j, z_j, r_j, g_j, b_j]^T$, where \( [x_j, y_j, z_j]^T \) are 3D coordinates and \( [r_j, g_j, b_j]^T \) are RGB values. 

Since a single 3D point can be visible from multiple views, it is straightforward to say that there can be multiple pixels from different images correspond to the same 3D point. We establish 2D-3D correspondences by projecting the sparse point cloud back onto the image plane of each view. The matching is defined as:

\begin{equation}
\mathbf{P}_j = f(\mathbf{V}_{i_1}, \mathbf{V}_{i_2}, \dots, \mathbf{V}_{i_m}).
\end{equation}
where \( \mathbf{V}_{i_k} \) are the pixels across different views \(k=1,\cdots, m\) that project to the same point \( \mathbf{P}_j \). Using \( f \), we evaluate the contribution of RGB images $\mathcal{I}=\{I_i\}_{i=1}^n$ to point clouds \( \mathcal{P}=\{\mathbf{P}_j\}_{j=1}^N \) based on the number pixel-to-point correspondences. The RGB image holds the most correspondences will be selected as the key frame for GP training. 

\textbf{GP Formulation.} Let the input features of the GP be denoted as \( \mathbf{X}_i = [u_i, v_i]^T \in \mathbb{R}^2 \), and the corresponding output targets as \( \mathbf{Y}_i = [x_i, y_i, z_i, r_i, g_i, b_i]^T \in \mathbb{R}^6 \). For notational convenience, we use \( \mathbf{x} := \mathbf{X}_i \in \mathbb{R}^2 \) and \( \mathbf{y} := \mathbf{Y}_i \in \mathbb{R}^6 \) to denote a single input-output pair. We define a multi-output GP over the function \( \mathbf{y}(\cdot): \mathbb{R}^2 \rightarrow \mathbb{R}^6 \) as:
\begin{equation}
    \mathbf{y}(\mathbf{x}) \sim \mathcal{GP}\left( \mathbf{m}(\mathbf{x}), \mathbf{K}(\mathbf{x}, \mathbf{x}') \right),
\end{equation}
where each output dimension corresponds to a coordinate in 3D space (position) or a colour channel (RGB). In practice, we employ independent kernels for each output dimension to capture their distinct spatial correlations, while predictive uncertainty is evaluated independently per channel to guide the densification process. Accordingly, the mean function \( \mathbf{m}(\mathbf{x}) \in \mathbb{R}^6 \) is defined as:
\begin{equation}
    \mathbf{m}(\mathbf{x}) = 
    \begin{bmatrix}
        m_1(\mathbf{x}) \\
        m_2(\mathbf{x}) \\
        \vdots \\
        m_6(\mathbf{x})
    \end{bmatrix},
    \quad \text{with } m_j: \mathbb{R}^2 \rightarrow \mathbb{R},
\end{equation}
and the covariance function \( \mathbf{K}(\mathbf{x}, \mathbf{x}') \in \mathbb{R}^{6 \times 6} \) is given by:
\begin{equation}
    \mathbf{K}(\mathbf{x}, \mathbf{x}') = 
    \begin{bmatrix}
        k_{11}(\mathbf{x}, \mathbf{x}') & \cdots & k_{16}(\mathbf{x}, \mathbf{x}') \\
        \vdots & \ddots & \vdots \\
        k_{61}(\mathbf{x}, \mathbf{x}') & \cdots & k_{66}(\mathbf{x}, \mathbf{x}')
    \end{bmatrix},
\end{equation}
\noindent
with $k_{ij}: \mathbb{R}^2 \times \mathbb{R}^2 \rightarrow \mathbb{R}$.

\textbf{Loss Function and Optimisation.} We train our GP model using:
\begin{equation}\label{eq:total-loss}
\mathcal{L}_{\text{total}} = -\log p(\mathbf{Y} | \mathbf{X}) + \lambda\|\boldsymbol{\theta}\|_2^2,
\end{equation}
where the first term maximises marginal likelihood and the second term is L2 regularisation to prevent overfitting.

For sparse SfM data, we employ the Matérn kernel rather than standard RBF, it provides a smoothness parameter \(\nu\) that balances accuracy and computational efficiency:
\begin{equation}\label{eq:matern-kernel}
    k_{\nu}(\mathbf{x}, \mathbf{x}') = \sigma^{2} \frac{2^{\,1-\nu}}{\Gamma(\nu)} \bigl(\sqrt{2\nu}\,\|\mathbf{x} - \mathbf{x}'\|\bigr)^{\nu} K_{\nu}\!\Bigl(\sqrt{2\nu}\,\|\mathbf{x} - \mathbf{x}'\|\Bigr),
\end{equation}
where smaller $\nu$ values allow sharper transitions whilst larger values enforce smoother reconstructions.

\subsection{Point Clouds Densification}
\label{Point Cloud Densification}

\begin{figure}[htbp]
    \centering    \includegraphics[width=1\linewidth]{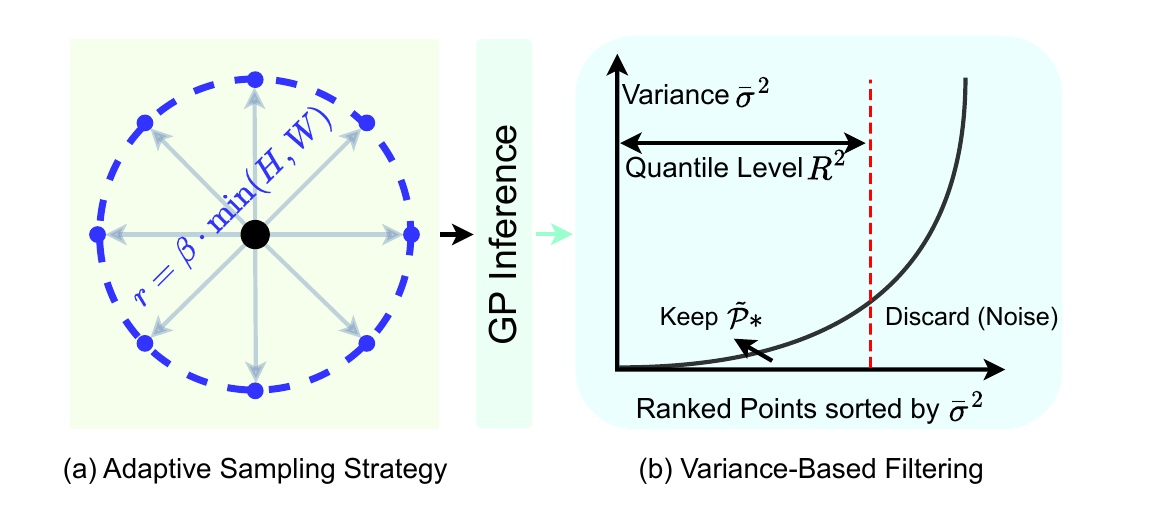}
 \caption{\textbf{Uncertainty-aware point clouds densification pipeline.} (a) Candidate pixels are sampled within a dynamic circular neighbourhood ($r = \beta \cdot \min(H, W)$) of valid points $(u_i, v_i)$ for GP inference. (b) Inferred points are ranked by variance $\bar{\sigma}^2$. A dynamic threshold $\tau$ (determined by quantile $R^2$) removes high-uncertainty artefacts.}
    
    \label{fig:morefiltered_variance}
\end{figure}

\textbf{Adaptive Sampling Strategy.} Figure~\ref{fig:morefiltered_variance} (a) illustrates our sampling strategy. To ensure local geometric consistency, we introduce an adaptive neighbourhood-based sampling mechanism that samples from immediate circular neighbourhoods of each available training point \((u_i, v_i)\). Specifically, we define a set of $N$ sampled pixels \(\mathcal{\tilde{P}}\) as follows:
\begin{align}
\mathcal{\tilde{P}}
=\;& \bigcup_{i=1}^N \bigcup_{j=1}^M 
\Bigl\{
\Bigl(\frac{u_i + r \cos \theta_j}{W},\, \frac{v_i + r \sin \theta_j}{H}\Bigr)\nonumber\\
&\quad \Bigm|\,
(u_i,\, v_i) \in \mathcal{V}_{i}
\Bigr\},
\label{eq:dynamic_sampling}
\end{align}
where 
$\theta_j \in [0, 2\pi)$ are uniformly distributed angles that control the sampling directions, $r = \beta \cdot \min(H, W)$ is the adaptive movement radius,  $\beta \in (0, 1)$ controls the sampling scale, \(W\) and \(H\) denote the image width and height, respectively. 
$W$ and $H$ are used to normalise all samples to the range \([0,1]\) in Equation (\ref{eq:dynamic_sampling}). The parameter $M$ determines the angular resolution of the sampling process.

\textbf{Variance-Based Filtering.}
The sampled pixels $\mathcal{\tilde{P}}$ are fed into the trained GP to predict dense points $\widehat{\mathcal{P}}$ and predictive variances $\mathbf{\Sigma} \in \mathbb{R}^{\widehat{N} \times 6}$. We identify unreliable predictions using the average variance across RGB channels, denoted as $\bar{\sigma}_i^2$, as this metric correlates strongly with visual artifacts. To filter these outliers, we rank the inferred points by $\bar{\sigma}_i^2$ and retain only the top quantile specified by the threshold $R^2$~\cite{edwards2008r2}. The cutoff value is defined as $\tau = \bar{\sigma}^2_{\lceil R^2 \cdot \widehat{N} \rceil}$, where $\lceil \cdot \rceil$ denotes the ceiling function. Predictions exceeding this uncertainty threshold are discarded to maintain geometric consistency, yielding a refined set of reliable points denoted as $\tilde{\mathcal{P}}^*$. Finally, the complete densified point cloud is formed by combining the original SfM points $\mathcal{P}$ with these filtered predictions: $\mathcal{P}_{Dense} = \mathcal{P} \cup \tilde{\mathcal{P}}^*$.

\section{Experiments}
\begin{figure}[t!]
    \centering
	\includegraphics[width=0.5\textwidth]{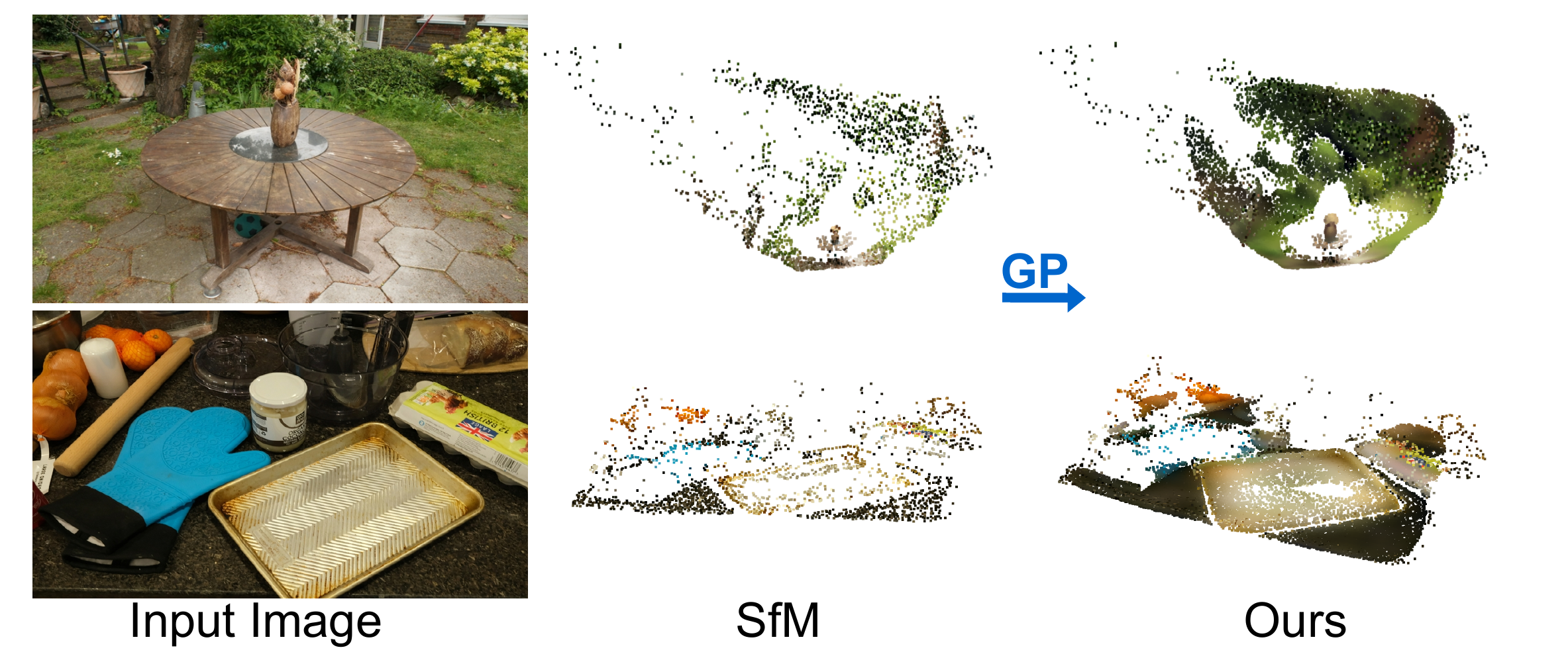}
    \caption{Visualisation of point cloud density on Mip-NeRF 360 (\textit{Garden} and \textit{Counter}).}
    \label{fig:gpr}
\end{figure}
\begin{figure}[t!]
    \centering
	\includegraphics[width=0.5\textwidth]{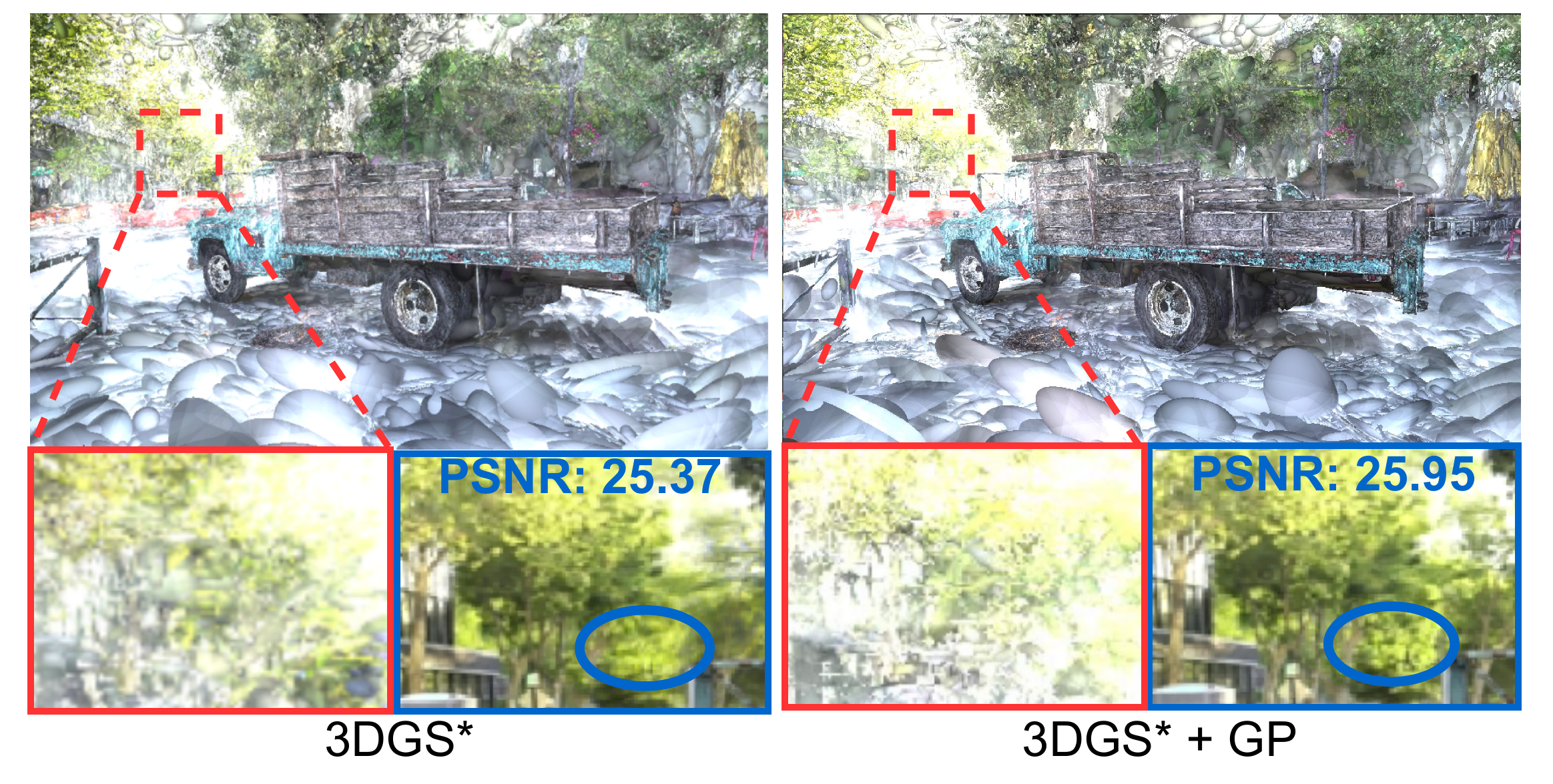}
    \caption{Qualitative comparison of 3D Gaussians during training process. The left column shows 3D Gaussians from 3DGS*, while the right column shows 3D Gaussians generated using ours GP.}
    \label{fig:3dgaussians}
\end{figure}
\begin{figure*}[t!]
    \centering
    \includegraphics[width=1\linewidth]{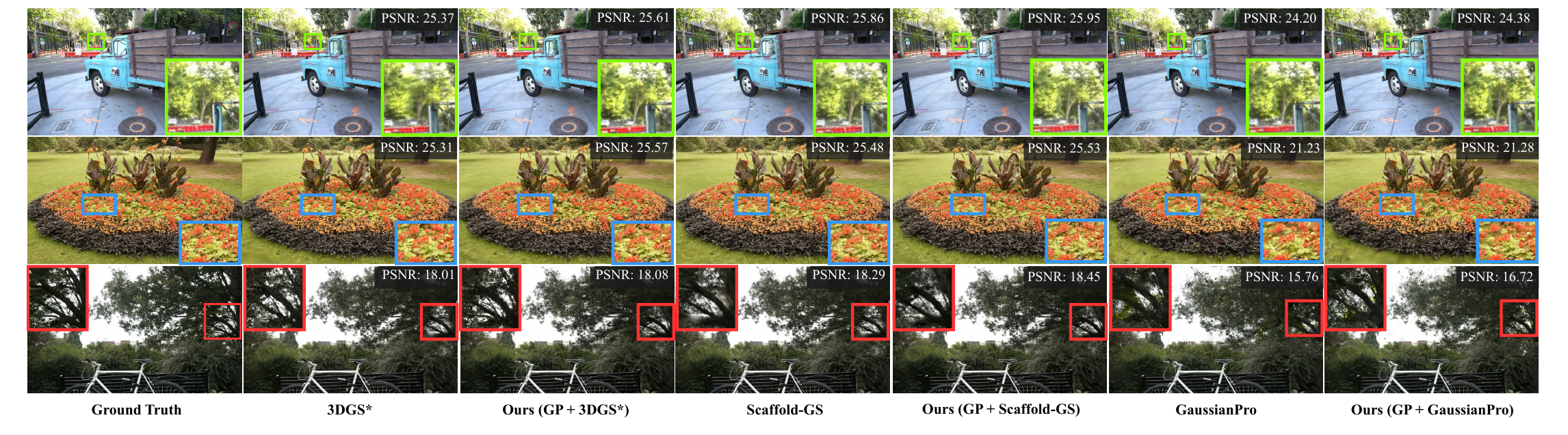}
    \caption{\textbf{Qualitative comparisons of novel view synthesis.} Rows from top to bottom display the \textit{Truck}, \textit{Flowers}, and \textit{Bicycle} scenes. Our method consistently recovers finer details (e.g., leaves, spokes, and edges) and preserves global geometry better across all scenes. These results further validate the effectiveness of our proposed plug-and-play GP module, which improves both 3DGS, Scaffold-GS and GaussianPro.
}
    \label{fig: comparison_quality}
\end{figure*}

\subsection{Dataset and Implementation Details}
\label{Dataset and Implementation Details}
\textbf{Dataset} Although SfM provides raw 2D-3D relationships, they are discrete and sparse. We explicitly organise these correspondences into a Pixel-to-Point dataset, transforming them from simple initialisation points into training data for continuous GP regression. We validate our method on standard benchmarks: NeRF Synthetic~\cite{mildenhall2021nerf}, Mip-NeRF 360~\cite{barron2022mip}, and Tanks \& Temples~\cite{knapitsch2017tanks}. Crucially, to ensure a strictly controlled comparison across all baselines and account for computational constraints, we standardised the evaluation resolutions: 800$\times$800 for NeRF Synthetic, $8\times$ downsampling for Mip-NeRF 360, and approx. 980$\times$545 for Tanks \& Temples. 

\textbf{Implementation Details} We incorporate GP into 3DGS* (officially improvement)~\cite{kerbl20233d}, Scaffold-GS~\cite{lu2024scaffold} and GaussianPro~\cite{cheng2024gaussianpro}. All experiments strictly follow the baseline's training/testing splits, hyperparameters and evaluation protocols. Our GP models are trained for 1,000 iterations across all scenes, we set L2 regularisation weight \(\lambda = 10^{-6}\), learning rate $l = 0.01$, dynamic sampling resolution $M = 8$, adaptive sampling radius $r = 0.25$. All experiments are conducted on an RTX 4080 GPU.

\textbf{Metrics.} 
For point cloud densification, we use Chamfer Distance (CD), 
Root Mean Squared Error (RMSE), and $R^2$ score~\cite{edwards2008r2}, 
following standard point cloud evaluation protocols~\cite{yu2021pointr}. For novel view rendering evaluation, we compare our method against state-of-the-art NVS approaches based on commonly used image-based metrics: Peak Signal-to-Noise Ratio (PSNR), Structural Similarity Index Measure (SSIM)~\cite{wang2004image}, and Learned Perceptual Image Patch Similarity (LPIPS)~\cite{zhang2018unreasonable} on the rendered images in the test views.
\begin{table}[t]
\centering
\footnotesize
\caption{Matérn kernel smoothness parameter study. Average metrics 
across all datasets (NeRF Synthetic, Tanks \& Temples, Mip-NeRF 360).}
\label{tab:matern-nu-study}
\begin{tabular}{lccc}
\toprule
\multicolumn{1}{c}{\textbf{\(\nu\)} Value} & \(R^2 \uparrow\) & \(\mathrm{RMSE} \downarrow\) & \(\mathrm{CD} \downarrow\) \\
\midrule
\rowcolor[gray]{0.8} 0.5 & \textbf{0.71} & \textbf{0.13} & \textbf{0.21} \\
1.5  & 0.67 & 0.15 & 0.24 \\
2.5  & 0.61 & 0.14 & 0.26 \\
\bottomrule
\end{tabular}
\vspace{-0.5cm}
\end{table}

\begin{table}[t]
\centering
\footnotesize
\caption{Inference Variance Analysis. We analyse the photometric variance ($\times 10^{-3}$) of the densified points inferred by our GP.}
\label{tab:dataset_variance_summary_averaged}
\setlength{\tabcolsep}{6pt}
\begin{tabular}{lccc}
\toprule
\textbf{Dataset} & \textbf{Original} & \textbf{Filtered} & \textbf{Reduction (\%)} \\
\midrule
Mip-NeRF 360      & 6.66   & 5.19   & 22.06 \\
Tanks \& Temples  & 32.31  & 22.65  & 29.90 \\
NeRF Synthetic    & 143.46 & 118.45 & 17.44 \\
\bottomrule
\end{tabular}
\end{table}

\begin{table*}[h!]
\centering
\caption{\textbf{Per-Scene Quantitative Comparison.} We use gray banding to distinguish baselines (Light Gray) from our method (Darker Gray). Metrics are reported row-wise as \textbf{PSNR $\uparrow$ SSIM $\uparrow$ LPIPS $\downarrow$}. Best results are \textbf{bold}.}
\label{tab:full_results_gray_clean}
\setlength{\tabcolsep}{3.0pt}
\small
\resizebox{\textwidth}{!}{%
\begin{tabular}{l|
>{\columncolor[gray]{0.9}}c | >{\columncolor[gray]{0.8}}c ||
>{\columncolor[gray]{0.9}}c | >{\columncolor[gray]{0.8}}c ||
>{\columncolor[gray]{0.9}}c | >{\columncolor[gray]{0.8}}c }
\toprule
\multirow{2}{*}{\textbf{Scene}} & 
\cellcolor{white}\textbf{GaussianPro} & \cellcolor{white}\textbf{Ours (+GP)} & 
\cellcolor{white}\textbf{Scaffold-GS} & \cellcolor{white}\textbf{Ours (+GP)} & 
\cellcolor{white}\textbf{3DGS} & \cellcolor{white}\textbf{Ours (+GP)} \\
& \cellcolor{white}PSNR SSIM LPIPS & \cellcolor{white}PSNR SSIM LPIPS& \cellcolor{white}PSNR SSIM LPIPS & \cellcolor{white}PSNR SSIM LPIPS & \cellcolor{white}PSNR SSIM LPIPS& \cellcolor{white}PSNR SSIM LPIPS \\
\midrule
\multicolumn{7}{c}{\cellcolor[gray]{0.95}\textit{\textbf{Mip-NeRF 360}}} \\
\midrule
Bicycle & 16.87 \ 0.473 \ 0.424 &17.60 \ 0.567 \ 0.362 & 19.70 \ 0.643 \ 0.262 & \textbf{19.80} \ \textbf{0.644} \ 0.260 & 19.17 \ 0.632 \ \textbf{0.248} & 19.22 \ 0.634 \ \textbf{0.248} \\
Bonsai & 17.87 \ 0.560 \ 0.476 & 17.90 \ 0.560 \ 0.471 & 18.25 \ 0.588 \ 0.445 & 18.50 \ \textbf{0.593 \ 0.437} & 18.66 \ 0.579 \ 0.458 & \textbf{18.69} \ 0.581 \ 0.455 \\
Counter & 27.97 \ 0.890 \ 0.150 & 28.13 \ 0.897 \ 0.150 & 29.73 \ 0.917 \ 0.121 & \textbf{29.83 \ 0.918 \ 0.119} & 29.15 \ 0.911 \ 0.122 & 29.17 \ 0.915 \ 0.121 \\
Garden & 27.62 \ 0.878 \ 0.100 & 27.66 \ 0.880 \ 0.099 & \textbf{29.56} \ 0.919 \ 0.064 & 29.53 \ 0.919 \ 0.064 & 29.10 \ 0.915 \ 0.063 & 29.38 \ \textbf{0.920 \ 0.061} \\
Kitchen & 29.89 \ 0.925 \ 0.092 & 30.01 \ 0.926 \ 0.092 & 31.86 \ 0.943 \ 0.073 & 31.87 \ 0.944 \ 0.073 & 31.74 \ 0.943 \ 0.069 & \textbf{32.28 \ 0.948 \ 0.068} \\
Room & 28.82 \ 0.899 \ 0.180 & 28.92 \ 0.896 \ 0.183 & 32.22 \ \textbf{0.939 \ 0.118} & \textbf{32.30 \ 0.939} \ 0.120 & 31.78 \ 0.934 \ 0.126 & 31.89 \ \textbf{0.939} \ 0.121 \\
Flowers & 21.11 \ 0.620 \ 0.297 & 21.22 \ 0.621 \ 0.297 & 23.08 \ 0.707 \ 0.237 & 23.15 \ 0.709 \ \textbf{0.236} & 23.04 \ 0.707 \ 0.251 & \textbf{23.16 \ 0.711} \ 0.246 \\
\textbf{Average} & 24.31 \ 0.749 \ 0.246 & 25.43 \ 0.779 \ 0.229 & 26.34 \ 0.808 \ 0.189 & \textbf{26.43 \ 0.809 \ 0.187} & 26.09 \ 0.803 \ 0.191 & 26.26 \ 0.807 \ 0.189 \\
\midrule
\multicolumn{7}{c}{\cellcolor[gray]{0.95}\textit{\textbf{Tanks \& Temples}}} \\
\midrule
Truck & 23.85 \ 0.846 \ 0.201 & 23.96 \ 0.846 \ 0.202 & 25.45 \ 0.879 \ 0.150 & \textbf{25.57 \ 0.880 \ 0.150} & 25.10 \ 0.876 \ 0.153 & 25.18 \ 0.877 \ 0.153 \\
Train & 20.49 \ 0.774 \ 0.262 & 20.73 \ 0.776 \ 0.261 & 22.50 \ 0.828 \ 0.198 & \textbf{22.53 \ 0.830 \ 0.196} & 22.13 \ 0.819 \ 0.202 & 22.22 \ 0.823 \ 0.199 \\
\textbf{Average} & 22.17 \ 0.810 \ 0.232 & 22.35 \ 0.811 \ 0.232 & 23.98 \ 0.854 \ 0.174 & \textbf{24.05 \ 0.855 \ 0.173} & 23.62 \ 0.848 \ 0.178 & 23.70 \ 0.850 \ 0.176 \\
\midrule
\multicolumn{7}{c}{\cellcolor[gray]{0.95}\textit{\textbf{NeRF Synthetic}}} \\
\midrule
Chair & 30.72 \ 0.972\ 0.022 & 30.84 \ 0.978 \ 0.021& 30.89 \ 0.976 \ 0.027 & 30.92 \ 0.976 \ 0.027 & 30.99 \ 0.978 \ 0.024 & \textbf{30.99 \ 0.978 \ 0.024} \\
Drums & 25.29 \ 0.937 \ \textbf{0.060} & 25.32 \ 0.938 \ \textbf{0.060} & 25.51 \ 0.936 \ 0.068 & \textbf{25.53}  \ 0.937\ 0.068 & 25.52 \ 0.940 \ 0.065 & 25.52 \ \textbf{0.940} \ 0.065 \\
Ficus & 23.96\ 0.905 \ 0.079 & 24.07 \ 0.908 \ 0.077 & 25.98 \ 0.925 \ 0.061 & \textbf{25.99 \ 0.927 }\ 0.060 & 25.03 \ 0.923 \ 0.058 & 25.20 \ 0.923 \ \textbf{0.058} \\
Hotdog & 34.23 \ \textbf{0.979} \ \textbf{0.035} & 34.25 \ \textbf{0.979} \ \textbf{0.035}& 34.51 \ 0.978 \ 0.042 & \textbf{34.61} \ 0.978 \ 0.041 & 34.53 \ \textbf{0.979} \ 0.037 & 34.55 \ \textbf{0.979} \ 0.037 \\
Lego & 29.03 \ 0.935 \ 0.066 & \textbf{29.09}\ 0.935 \ \textbf{0.065} & 28.62 \ 0.926 \ 0.075 & 28.90 \ 0.929 \ 0.074 & 28.32 \ 0.929 \ 0.074 & 29.08 \ \textbf{0.936} \ 0.067 \\
Materials & 17.20 \ 0.752 \ 0.235& 17.30 \ \textbf{0.755 }\ 0.233& 17.88 \ 0.749 \ 0.221 & \textbf{18.11 \ 0.755 \ 0.219} & 16.82 \ 0.734 \ 0.239 & 16.68 \ 0.736 \ 0.241 \\
Mic & 21.05 \ 0.868 \ 0.136 & 21.61 \ \textbf{0.876 }\ 0.127 & 27.67 \ 0.852 \ 0.153 & \textbf{27.71} \ 0.853 \ 0.152 & 18.90 \ 0.855 \ 0.144 & 19.02 \ 0.863 \ \textbf{0.123 }\\
Ship & 26.57 \ 0.843 \ 0.143 & 26.61 \ 0.843 \ \textbf{0.142} & \textbf{27.66 \ 0.853} \ 0.153 & 27.65 \ 0.852 \ 0.151 & 27.08 \ 0.848 \ 0.157 & 27.05 \ 0.849 \ 0.154 \\
\textbf{Average} & 26.01 \ 0.899 \ 0.097 & 26.14 \ \textbf{0.902} \ \textbf{0.095} & 27.34 \ 0.899 \ 0.100 & \textbf{27.43} \ 0.901 \ 0.099 & 25.90 \ 0.898 \ 0.100 & 26.01 \ 0.901 \ 0.096 \\
\bottomrule
\end{tabular}
}
\end{table*}

\subsection{Results of Gaussian Process}
\textbf{Quantitative Results} We evaluated the GP model on our Pixel-to-Point dataset using an 80/20 train-test split and varying smoothness parameters ($\nu \in \{0.5, 1.5, 2.5\}$). As summarised in Table~\ref{tab:matern-nu-study}, $\nu = 0.5$ consistently yielded the best performance across $R^2$, RMSE, and CD metrics, we therefore adopt it as the default configuration. Detailed metrics across multiple datasets for this kernel selection procedure are presented in \textbf{Appendix Table~\ref{tab:PerformanceMetrics}}. Table~\ref{tab:dataset_variance_summary_averaged} compares the average predictive variance of our raw GP predictions against the subset retained after variance-based pruning. (\textbf{As detailed in Table~\ref{tab:full_rgb_variance_averages_consistent}}). For real-world datasets like Tanks \& Temples and Mip-NeRF 360, the high reduction percentages (29.90\% and 22.06\%) confirm that our GP correctly captures the inherent uncertainty present in complex outdoor scenes, assigning high variance to unreliable regions which are then effectively filtered.

\textbf{Qualitative Results} 
Figure~\ref{fig:gpr} presents the Visualisation of point cloud density on Mip-NeRF 360 (\textit{Garden} and \textit{Counter}). Standard SfM struggles with sparsity in smooth regions, whereas our GP based desification method recovers dense, geometrically consistent surfaces. Note the improved surface continuity on the table and the restoration of fine structural details in the foliage. Furthermore, Figure~\ref{fig:3dgaussians} visualises the structural benefits of our method. While the baseline 3DGS* struggles to reconstruct dense foliage, resulting in gaps and blur, our GP-enhanced initialisation ensures a complete and consistent scene representation. This is quantitatively supported by the observed PSNR improvement of 0.58 dB.

\subsection{Quantitative and Qualitative Results on NVS}
\label{Quantitative and Qualitative Results}

\textbf{Quantitative Results} As detailed in Table~\ref{tab:full_results_gray_clean}, our method consistently outperforms state-of-the-art baselines. On the NeRF Synthetic dataset, we achieve an average PSNR of 27.43 dB, surpassing Scaffold-GS by 0.09 dB and 3DGS by 1.53 dB. Notable improvements are observed in challenging scenes such as Materials, where our method boosts PSNR by 0.23 dB (17.88 $\rightarrow$ 18.11) compared to Scaffold-GS, indicating superior handling of complex surface reflections. Similarly, on real-world datasets like Mip-NeRF 360 and Tanks \& Temples, our approach maintains robust performance. We achieve the highest average PSNR of 26.43 dB and 24.05 dB, respectively, consistently yielding lower LPIPS scores. This confirms that our uncertainty-guided densification not only improves rendering fidelity but also aligns better with human perceptual quality.

\textbf{Qualitative Results} Figure~\ref{fig: comparison_quality} visualises the improvements of our GP module when integrated into 3DGS*, Scaffold-GS and GaussianPro. Our method consistently restores fine-grained details often lost in baseline reconstructions. In the \textit{Truck} and \textit{Flowers} scenes, our approach sharply recovers the intricate textures of foliage and flowers that appear blurred in the baselines. Similarly, in the challenging \textit{Bicycle} scene, GP-GS successfully reconstructs thin structures like branches, reducing aliasing and geometric gaps. These results validate GP-GS as a robust plug-and-play module that enhances perceptual quality and structural consistency in complex environments.

\section{Ablation Study}
\begin{table}[htbp]
\centering
\caption{The impact of kernel choice and input features on inference accuracy, averaged over the Tanks \& Temples dataset (Truck and Train).}
\label{tab:gp_ablation}
\setlength{\tabcolsep}{5pt}
\begin{tabular}{llccc}
\toprule
Input Features & Kernel Type & R$^2$ Score & RMSE & CD \\
\midrule
XY + Depth & Matern ($\nu=1/2$) & 0.625 & \textbf{0.153} & \textbf{0.115} \\
XY Only & Matern ($\nu=1/2$) & \textbf{0.626} & \textbf{0.153} & \textbf{0.115} \\
\midrule
XY + Depth & RBF  & 0.602 & 0.159 & 0.192 \\
XY Only & RBF  & 0.602 & 0.159 & 0.191 \\
\bottomrule
\end{tabular}
\end{table}
This section further investigates how GP kernel selections and depth priors affect the performance of the proposed method. Table~\ref{tab:gp_ablation} evaluates kernel and input features selection on \textit{Tanks \& Temples} dataset. We find that the Matern kernel ($\nu=0.5$) significantly outperforms RBF kernel (CD $0.115$ vs. $0.191$), proving effective for modelling discontinuous SfM data. Furthermore, incorporating depth priors~\cite{yang2024depth} yields no accuracy gain over 2D coordinates alone, confirming that 2D spatial distribution suffices for local inference and justifying our lightweight, SfM-only design.

\section{Conclusion}
\label{conclusion}

We presented GP-GS, a Gaussian-Process-based approach for uncertainty-aware densification of sparse SfM point clouds to improve 3DGS. By formulating densification as a local regression problem constrained by existing geometry, our method interpolates additional 3D points while explicitly modelling predictive uncertainty. An adaptive neighbourhood-based sampling strategy and variance-based filtering mechanism reduce noise and preserve geometric structure. Extensive experiments demonstrate that GP-GS consistently improves rendering fidelity across synthetic and real-world datasets, and can be seamlessly integrated into existing 3DGS pipelines.

\bibliographystyle{IEEEbib}
\bibliography{gpgs}

\clearpage 

\onecolumn
\section{GP Details}
\subsection{GP Introduction}
Mathematically, a GP is fully specified by its mean function $m(\mathbf{x})$ and covariance function $k\left(\mathbf{x}, \mathbf{x}^{\prime}\right)$, a.k.a. kernel function, where $\mathbf{x}$ and $\mathbf{x}^{\prime}$ are inputs. The kernel function typically depends on a set of hyperparameters  $\boldsymbol{\theta}$, such as length scale \( l \), signal variance \( \sigma_f^2 \), or other parameters depending on the kernel type (e.g., Matérn). A GP is usually used as a prior over a latent function defined as:
\begin{equation}
    f(\mathbf{x}) \sim \mathcal{GP}\Big(m(\mathbf{x}), k(\mathbf{x}, \mathbf{x}')\Big). 
\end{equation}
Training a GP model involves optimizing the hyperparameters \( \boldsymbol{\theta} \) of the kernel function to maximize the likelihood of the observed data. This is achieved by maximizing the log marginal likelihood, given by:
\begin{equation}
\log p(\mathbf{y} \mid \mathbf{X}, \boldsymbol{\theta})=-\frac{1}{2} \mathbf{y}^T \mathbf{K}^{-1} \mathbf{y}-\frac{1}{2} \log |\mathbf{K}|-\frac{n}{2} \log 2 \pi,
\end{equation}
\noindent
where $\mathbf{y}$ are the observed outputs, $\mathbf{X}$ are the observed inputs, \( \mathbf{K} \in \mathbb{R}^{n \times n} \) is the Gram matrix with entries computed using the kernel function \( k(\mathbf{x}, \mathbf{x}^\prime) \) across all training inputs, and \( n \) is the number of training points. After training, the predictive distribution at a new test point \( \mathbf{x}_* \) follows a Gaussian distribution: $f\left(\mathbf{x}_*\right) \mid \mathbf{X}, \mathbf{y}, \mathbf{x}_* \sim \mathcal{N}\left(\mu_*, \sigma_*^2\right)$, where the predictive mean and variance are computed as:
\begin{equation}
    \mu_* = \mathbf{k}_*^T \mathbf{K}^{-1} \mathbf{y}, \quad
    \sigma_*^2 = k(\mathbf{x}_*, \mathbf{x}_*) - \mathbf{k}_*^T \mathbf{K}^{-1} \mathbf{k}_*,
\end{equation}
\noindent
where \( \mathbf{k}_* = [k(\mathbf{x}_*, \mathbf{x}_i)]^T, i=1, \cdots, n, \in \mathbb{R}^{n \times 1} \) represents the covariance between the test point \( \mathbf{x}_* \) and all training points, while \( k(\mathbf{x}_*, \mathbf{x}_*) \) denotes the variance of the test point itself.

\subsection{GP Training}
\textbf{GP Training Procedure} Given a set of $n$ RGB images $\mathcal{I}=\{I_i\}_{i=1}^n$ and the corresponding sparse point clouds \( \mathcal{P}=\{\mathbf{P}_j\}_{j=1}^N \) with a total number of \( N \) points. Our goal is to predict dense point clouds given image pixels as input. Our GP training procedure can be found in Algorithm~\ref{alg:mogp}.
\begin{algorithm}[th]
\caption{GP with Matérn Kernel}
\label{alg:mogp}

\textbf{Input:}  
RGB images \(\mathcal{I}\), sparse 3D points \(\mathcal{P}\), pixel correspondences \(\mathcal{V}\).  
Feature inputs \(\mathbf{X} \in \mathbb{R}^{n \times 2}\) , target outputs \(\mathbf{Y} \in \mathbb{R}^{n \times 6}\).

\textbf{Step 1: GP Definition}  
\begin{equation}
\mathbf{y} \sim \mathcal{GP}(\mathbf{m}(\mathbf{x}), \mathbf{K}(\mathbf{x}, \mathbf{x}'))
\end{equation}
where \(\mathbf{m}(\mathbf{x})\) is the mean function and \(\mathbf{K}(\mathbf{x}, \mathbf{x}')\) is the covariance function.

\textbf{Step 2: Matérn Kernel}
\begin{equation}
k_{\nu}(\mathbf{x}, \mathbf{x}') = \sigma^{2} \frac{2^{1-\nu}}{\Gamma(\nu)}
(\sqrt{2\nu} \|\mathbf{x} - \mathbf{x}'\|)^{\nu} K_{\nu}(\sqrt{2\nu} \|\mathbf{x} - \mathbf{x}'\|)
\end{equation}
where \(\nu\) controls smoothness and \(\sigma^2\) defines variance.

\textbf{Step 3: Optimisation}  
\textbf{Loss Function:}  
\begin{equation}
\mathcal{L}_{\text{total}} = -\log p(\mathbf{Y} | \mathbf{X}) + \lambda\|\boldsymbol{\theta}\|_2^2 
\end{equation}
where \(\lambda = 10^{-6}\) is the L2 regularisation weight.

\textbf{Gradient-Based Parameter Update:}  
\begin{itemize}
    \item Initialize: $\boldsymbol{\theta} = [\sigma, \nu, \dots]^T$. 
    \item \textbf{for} \(t = 1 \dots T\):
    \begin{itemize}
        \item Compute gradient \(\nabla_{\boldsymbol{\theta}} \mathcal{L}_{\text{total}}\).
        \item Update:
        \begin{equation}
            \boldsymbol{\theta} \leftarrow \boldsymbol{\theta} - \eta \,\nabla_{\boldsymbol{\theta}}\,\mathcal{L}_{\text{total}}
        \end{equation}
    \end{itemize}
\end{itemize}

\textbf{Output:}  
Optimised GP hyperparameters \(\theta\).  
Predicted outputs \(\widehat{\mathbf{Y}}\) for new inputs \(\widehat{\mathbf{X}}\) via the GP posterior.

\end{algorithm}

\textbf{Smoothness Parameters \(\nu\) Selection}
We employ the Matérn kernel for training our GP model. In contrast to the standard RBF kernel, the Matérn kernel introduces a smoothness parameter \(\nu\), which governs the differentiability and local variations of the latent function, and ends up with a better trade-off between smoothness and computational efficiency. Formally, the Matérn kernel between two points \(\mathbf{x}\) and \(\mathbf{x}'\) is defined as:
\begin{equation}\label{eq:matern-kernel}
    k_{\nu}(\mathbf{x}, \mathbf{x}') \;=\; 
    \sigma^{2}
    \frac{2^{\,1-\nu}}{\Gamma(\nu)} 
    \bigl(\sqrt{2\nu}\,\|\mathbf{x} - \mathbf{x}'\|\bigr)^{\nu}\,
    K_{\nu}\!\Bigl(\sqrt{2\nu}\,\|\mathbf{x} - \mathbf{x}'\|\Bigr),
\end{equation}
where \(\Gamma(\cdot)\) is the Gamma function, \(\sigma^2\) is the variance, and \(K_{\nu}(\cdot)\) is a modified Bessel function of the second kind. The parameter \(\nu\) modulates how smooth the function can be: smaller \(\nu\) values permit abrupt transitions, while larger \(\nu\) values enforce smoother spatial changes. Details of tuning \(\nu\) on various datasets can be found in Table~\ref{tab:PerformanceMetrics}.

\begin{table*}[htbp]
  \centering
  \footnotesize
  \caption{Performance Metrics for GP Model Using Different Matérn Kernel Parameters. The best results are highlighted in \textbf{bold}. The results indicate that $\nu = 0.5$ consistently produces the best densification accuracy across varying datasets.}
  \label{tab:PerformanceMetrics}
  \begin{tabular}{ll >{\columncolor{gray!15}}c >{\columncolor{gray!15}}c >{\columncolor{gray!15}}c cccccc}
        \toprule
        & & \multicolumn{3}{c}{\cellcolor{gray!15}$\nu = 0.5$} & \multicolumn{3}{c}{$\nu = 1.5$} & \multicolumn{3}{c}{$\nu = 2.5$} \\
        \cmidrule(r){3-5} \cmidrule(l){6-8} \cmidrule(l){9-11}
        Main Dataset & Sub Dataset & $R^2$ $\uparrow$ & RMSE $\downarrow$ & CD $\downarrow$ & $R^2$ $\uparrow$ & RMSE $\downarrow$ & CD $\downarrow$ & $R^2$ $\uparrow$ & RMSE $\downarrow$ & CD $\downarrow$ \\
        \midrule
        \multirow{8}{*}{\textbf{NeRF Synthetic}} 
        & Lego                & \textbf{0.78} & \textbf{0.12} & \textbf{0.18} & 0.75 & 0.13 & 0.20 & 0.69 & 0.15 & 0.21 \\
        & Chair               & \textbf{0.63} & \textbf{0.14} & \textbf{0.19} & 0.57 & 0.15 & 0.28 & 0.48 & 0.15 & 0.35 \\
        & Drums               & 0.52 & 0.22 & \textbf{0.61} & 0.52 & 0.22 & 0.64 & 0.52 & 0.22 & 0.64 \\
        & Hotdog              & \textbf{0.65} & \textbf{0.09} & \textbf{0.16} & 0.52 & 0.51 & 0.43 & 0.38 & 0.12 & 0.22 \\
        & Ficus              & \textbf{0.52} & \textbf{0.22} & \textbf{0.68} & 0.46 & 0.23 & 0.77 & 0.45 & 0.23 & 0.77 
        \\
        & Materials              & \textbf{0.53} & \textbf{0.19} & \textbf{0.24} & 0.49 & 0.20 & 0.24 & 0.48 & 0.20 & 0.39 
        \\
        & Ship              & \textbf{0.55} & \textbf{0.11} & \textbf{0.22} & 0.51 & 0.12 & 0.23 & 0.25 & 0.14 & 0.23 
        \\
        & Mic              & \textbf{0.45} & 0.23 & \textbf{0.69} & 0.42 & 0.23 & 0.70 & 0.43 & 0.23 & 0.71 
        \\
        & Average              & \textbf{0.58} & \textbf{0.17} & \textbf{0.13} & 0.53 & 0.22 & 0.44 & 0.46 & 0.18 & 0.44
        \\
        \midrule
        \multirow{2}{*}{\textbf{Tanks \& Temples}} 
        & Truck                & \textbf{0.78} & \textbf{0.12} & \textbf{0.18} & 0.75 & 0.13 & 0.20 & 0.69 & 0.15 & 0.21 \\
        & Train               & \textbf{0.78} & \textbf{0.10} & \textbf{0.05} & 0.74 & \textbf{0.10} & 0.07 & 0.73 & 0.11 & 0.07 
        \\
        & Average              & \textbf{0.78} & \textbf{0.11} & \textbf{0.11} & 0.75 & 0.12 & 0.14 & 0.71 & 0.13 & 0.14
        \\
        \midrule
        \multirow{8}{*}{\textbf{Mip-NeRF 360}} 
        & Bicycle                & \textbf{0.74} & \textbf{0.10} & \textbf{0.12} & 0.70 & 0.11 & 0.12 & 0.47 & 0.16 & 0.41 \\
        & Bonsai               & \textbf{0.77} & \textbf{0.12} & \textbf{0.19} & 0.72 & 0.13 & 0.23 & 0.70 & 0.13 & 0.25 \\
        & Counter               & \textbf{0.85} & \textbf{0.10} & \textbf{0.13} & 0.82 & 0.11 & \textbf{0.13} & 0.82 & 0.11 & 0.14 \\
        & Garden               & \textbf{0.66} & \textbf{0.12} & \textbf{0.13} & 0.60 & 0.13 & \textbf{0.13} & 0.57 & 0.13 & 0.14 \\
        & Kitchen               & \textbf{0.82} & \textbf{0.08} & \textbf{0.05} & 0.79 & \textbf{0.08} & 0.06 & 0.79 & 0.09 & 0.08 
        \\
        & Room               & \textbf{0.74} & \textbf{0.097} & \textbf{0.136} & 0.71 & 0.104 & 0.143 & 0.68 & 0.109 & 0.156 
        \\
        & Flowers               & \textbf{0.75} & \textbf{0.115} & \textbf{0.15} & 0.72 & 0.122 & 0.214 & 0.69 & 0.125 & 0.152
        \\
        & Average              & \textbf{0.77} & \textbf{0.10} & \textbf{0.13} & 0.72 & 0.11 & 0.15 & 0.67 & 0.12 & 0.19
        \\
        \bottomrule
    \end{tabular}
\end{table*}

\begin{figure}[htbp] 
    \centering
    \includegraphics[width=0.7\linewidth]
    {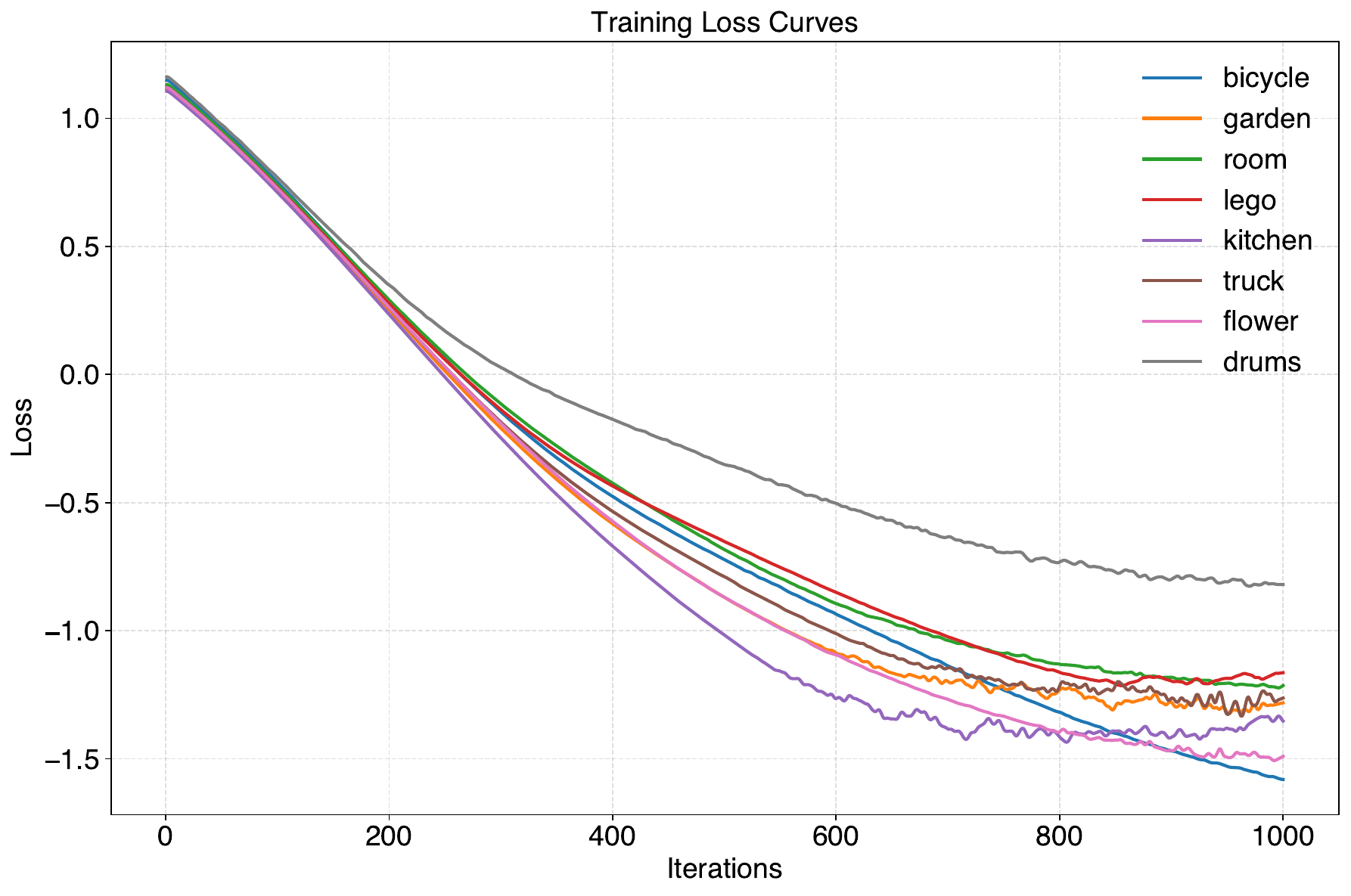} 
    \caption{GP training loss curves for different datasets over 1,000 iterations. The loss demonstrates consistent convergence as training progresses.}
    \label{fig:mogploss}
\end{figure}

\textbf{GP Training Loss.} We present the training loss curves of our GP model across varying datasets to illustrate convergence behaviour. Figure~\ref{fig:mogploss} depicts the loss curves over 1,000 iterations for scenes including Bicycle, Garden, Room, Lego, Kitchen, Truck, Flowers, and Drums. The curves demonstrate a smooth and consistent decline, indicating stable optimisation and effective learning of the geometric and colour attributes.

\textbf{GP Evaluation Metrics Details.} For the evaluation of the SfM sparse point densification, in line with prior studies on point cloud reconstruction and completion~\cite{yu2021pointr,yuan2018pcn}, we adopt the mean Chamfer Distance (CD) to measure the discrepancy between the predicted point clouds and the ground truth. Specifically, for a predicted point set $\mathcal{P}$ and its corresponding ground truth set $\mathcal{G}$, the CD between them is calculated as:
\begin{align}
d_{CD}(\mathcal{P}, \mathcal{G}) &= \frac{1}{|\mathcal{P}|} 
\sum_{\mathbf{p_i} \in \mathcal{P}} 
\min _{\mathbf{g_i} \in \mathcal{G}} \|\mathbf{p_i}-\mathbf{g_i}\| \nonumber \\
&\quad + \frac{1}{|\mathcal{G}|} 
\sum_{\mathbf{g_i} \in \mathcal{G}} 
\min _{\mathbf{p_i} \in \mathcal{P}} \|\mathbf{g_i}-\mathbf{p_i}\|.
\end{align}

We also compute the Root Mean Squared Error (RMSE) and $R^2$~\cite{edwards2008r2} to show the robustness of GP-GS under various metrics. The RMSE is calculated to quantify the average squared deviation between the predicted and true values, using $\mathrm{RMSE} = \sqrt{\frac{1}{n} \sum_{i=1}^{n} (\mathbf{g_i} - \mathbf{p_i})^2}$, a lower RMSE value indicates a more accurate prediction. The \(R^2\) score captures the proportionate reduction in the residual variance:
\begin{equation}
\label{r2}
    R^2 = 1 - \frac{\sum_{i=1}^{n} (\mathbf{g_i} - \mathbf{p_i})^2}{\sum_{i=1}^{n} (\mathbf{g_i} - \bar{\mathbf{g_i}})^2},
\end{equation}
\noindent
where \(\mathbf{g_i}\) is the true value,  \(\mathbf{p_i}\) is the predicted value, and $\bar{\mathbf{g_i}}$ is the mean of the true values. A higher \(R^2\) (closer to 1) indicates better Gaussian process model performance.

\subsection{GP Inference}
Ours GP inference include Adaptive Sampling and Variance-Based Filtering, an overview of the uncertainty-based filtering procedure can be found in Algorithm~\ref{alg:dynamic_sampling_filtering}.
\begin{algorithm}[ptbh]
\caption{Adaptive Sampling  and Variance-Based Filtering}
\label{alg:dynamic_sampling_filtering}

\textbf{Input:}  
Image size \((W, H)\), training pixel coordinates \(\{[u_i, v_i]^T\}_{i=1}^N\),  
angular resolution \(M\), scale factor \(\beta\), variance quantile threshold \(R^2\).  

\textbf{Step 1: Adaptive Sampling}  
\begin{enumerate}
  \item Compute adaptive sampling radius \(r = \beta \cdot \min(W, H)\).
  \item \textbf{for} each training pixel with coordinates \([u_i, v_i]^T\):
    \begin{itemize}
      \item Generate \(M\) new samples stacked as \([\mathbf{u}, \mathbf{v}]^T \in \mathbb{R}^{M\times 2}\) in a circular neighborhood.
      \item Store \(\bigl[\mathbf{u}, \mathbf{v}\bigr]^T\) in sample set \(\mathcal{\tilde{P}}\).
    \end{itemize}
\end{enumerate}

\textbf{Step 2: GP Inference}  
Provide \(\mathcal{\tilde{P}}\) to GP to infer a densified point cloud \(\widehat{\mathcal{P}}\) with variance matrix \(\mathbf{\Sigma}\).

\textbf{Step 3: Variance-Based Filtering}  
\begin{enumerate}
  \item Compute mean RGB variance \(\bar{\sigma}^2\) for each inferred point.
  \item Set threshold \(\tau\) as the \(R^2\)-quantile of the variance distribution.
  \item Remove points with \(\bar{\sigma}^2 > \tau\).
\end{enumerate}

\textbf{Step 4: Final Point Cloud}  
\(\mathcal{\tilde{P}}^* \gets \mathcal{P} \cup \widehat{\mathcal{P}}\).  

\textbf{Output:} Densified and filtered point clouds \(\mathcal{\tilde{P}}^*\).

\end{algorithm}

\textbf{Variance Analysis.} To validate the effectiveness of our variance-based filtering algorithm, we analyse the variance distributions across multiple Pixel-to-Point datasets, covering diverse characteristics from NeRF Synthetic, Tanks \& Temples, and Mip-NeRF 360. As detailed in Table~\ref{tab:full_rgb_variance_averages_consistent}, the significant reduction in variance across all datasets suggests that our filtering method generalises well to different 3D scene types. By effectively removing high-variance noisy points while preserving scene consistency, the method improves overall point cloud quality.
\begin{table*}[htbp]
\centering
\caption{\textbf{Detailed Photometric Variance Analysis.} We report the per-channel variance ($\times 10^{-3}$) before and after applying our uncertainty-based filtering across all datasets. The columns show the Original ($\sigma^2_{orig}$), Filtered ($\sigma^2_{filt}$), and the relative Reduction ($\Delta\%$) for Red, Green, and Blue channels. The method demonstrates robust performance across varying noise levels.}
\label{tab:full_rgb_variance_averages_consistent}
\setlength{\tabcolsep}{3.5pt}
\begin{tabular}{lc|ccc|ccc|ccc}
\toprule
 & \textbf{Filter} & \multicolumn{3}{c|}{\textbf{Original Variance} ($\times 10^{-3}$)} & \multicolumn{3}{c|}{\textbf{Filtered Variance} ($\times 10^{-3}$)} & \multicolumn{3}{c}{\textbf{Reduction} (\%)} \\
\cmidrule(lr){3-5} \cmidrule(lr){6-8} \cmidrule(lr){9-11}
\textbf{Dataset / Scene} & \textbf{Thres.} & \textbf{Red} & \textbf{Green} & \textbf{Blue} & \textbf{Red} & \textbf{Green} & \textbf{Blue} & \textbf{Red} & \textbf{Green} & \textbf{Blue} \\
\midrule

\multicolumn{11}{c}{\cellcolor[gray]{0.95}\textbf{Mip-NeRF 360}} \\
Bicycle & 74\% & 7.65 & 6.83 & 3.45 & 5.23 & 4.13 & 2.78 & 31.63 & 39.53 & 19.42 \\
Room & 74\% & 7.56 & 6.06 & 3.24 & 5.65 & 4.57 & 2.74 & 25.26 & 24.59 & 15.43 \\
Bonsai & 77\% & 23.72 & 16.88 & 31.34 & 18.12 & 14.22 & 24.00 & 23.61 & 15.76 & 23.42 \\
Counter & 85\% & 2.61 & 2.31 & 2.73 & 2.23 & 2.03 & 2.28 & 14.56 & 12.12 & 16.48 \\
Flowers & 75\% & 2.43 & 2.19 & 1.85 & 2.15 & 2.00 & 1.68 & 11.52 & 8.68 & 9.19 \\
Garden & 66\% & 3.70 & 3.99 & 2.61 & 3.00 & 3.22 & 2.18 & 18.92 & 19.30 & 16.48 \\
Kitchen & 82\% & 2.43 & 3.39 & 2.94 & 1.92 & 2.62 & 2.35 & 20.99 & 22.71 & 20.07 \\
\rowcolor[gray]{0.98} \textit{Average} & - & \textit{7.16} & \textit{5.95} & \textit{6.88} & \textit{5.47} & \textit{4.68} & \textit{5.43} & \textit{\textbf{23.60}} & \textit{\textbf{21.34}} & \textit{\textbf{21.08}} \\
\midrule

\multicolumn{11}{c}{\cellcolor[gray]{0.95}\textbf{Tanks \& Temples}} \\
Truck & 78\% & 51.30 & 47.27 & 78.65 & 36.08 & 34.15 & 55.27 & 29.67 & 27.76 & 29.73 \\
Train & 78\% & 5.49 & 5.71 & 5.43 & 3.65 & 3.34 & 3.40 & 33.52 & 41.51 & 37.38 \\
\rowcolor[gray]{0.98} \textit{Average} & - & \textit{28.40} & \textit{26.49} & \textit{42.04} & \textit{19.87} & \textit{18.75} & \textit{29.34} & \textit{\textbf{30.04}} & \textit{\textbf{29.22}} & \textit{\textbf{30.21}} \\
\midrule

\multicolumn{11}{c}{\cellcolor[gray]{0.95}\textbf{NeRF Synthetic}} \\
Chair & 63\% & 512.62 & 275.14 & 346.06 & 420.70 & 203.79 & 250.43 & 17.93 & 25.93 & 27.63 \\
Drums & 52\% & 80.12 & 93.61 & 85.78 & 79.15 & 93.00 & 85.49 & 1.21 & 0.65 & 0.34 \\
Ficus & 52\% & 92.82 & 101.75 & 67.83 & 91.14 & 95.62 & 65.27 & 1.81 & 6.02 & 3.77 \\
Hotdog & 65\% & 21.97 & 96.04 & 17.38 & 13.55 & 55.25 & 11.58 & 38.32 & 42.47 & 33.37 \\
Lego & 78\% & 394.56 & 346.51 & 233.23 & 288.57 & 249.70 & 178.14 & 26.86 & 27.94 & 23.62 \\
Materials & 53\% & 101.83 & 100.46 & 93.91 & 100.56 & 99.09 & 92.48 & 1.25 & 1.36 & 1.52 \\
Mic & 45\% & 108.21 & 86.88 & 90.17 & 103.50 & 83.60 & 87.74 & 4.35 & 3.78 & 2.69 \\
Ship & 55\% & 31.44 & 33.78 & 30.97 & 30.85 & 33.13 & 30.48 & 1.88 & 1.92 & 1.58 \\
\rowcolor[gray]{0.98} \textit{Average} & - & \textit{167.95} & \textit{141.77} & \textit{120.67} & \textit{141.00} & \textit{114.15} & \textit{100.20} & \textit{\textbf{16.05}} & \textit{\textbf{19.48}} & \textit{\textbf{16.96}} \\
\midrule

\rowcolor[gray]{0.8} \textbf{Global Avg} & - & \textbf{85.32} & \textbf{72.28} & \textbf{64.56} & \textbf{70.94} & \textbf{57.85} & \textbf{52.84} & \textbf{16.85} & \textbf{19.96} & \textbf{18.15} \\
\bottomrule
\end{tabular}
\end{table*}

\section{Enhanced 3DGS Details}

\begin{figure}[htbp]
    \centering
    \includegraphics[width=0.8\linewidth]{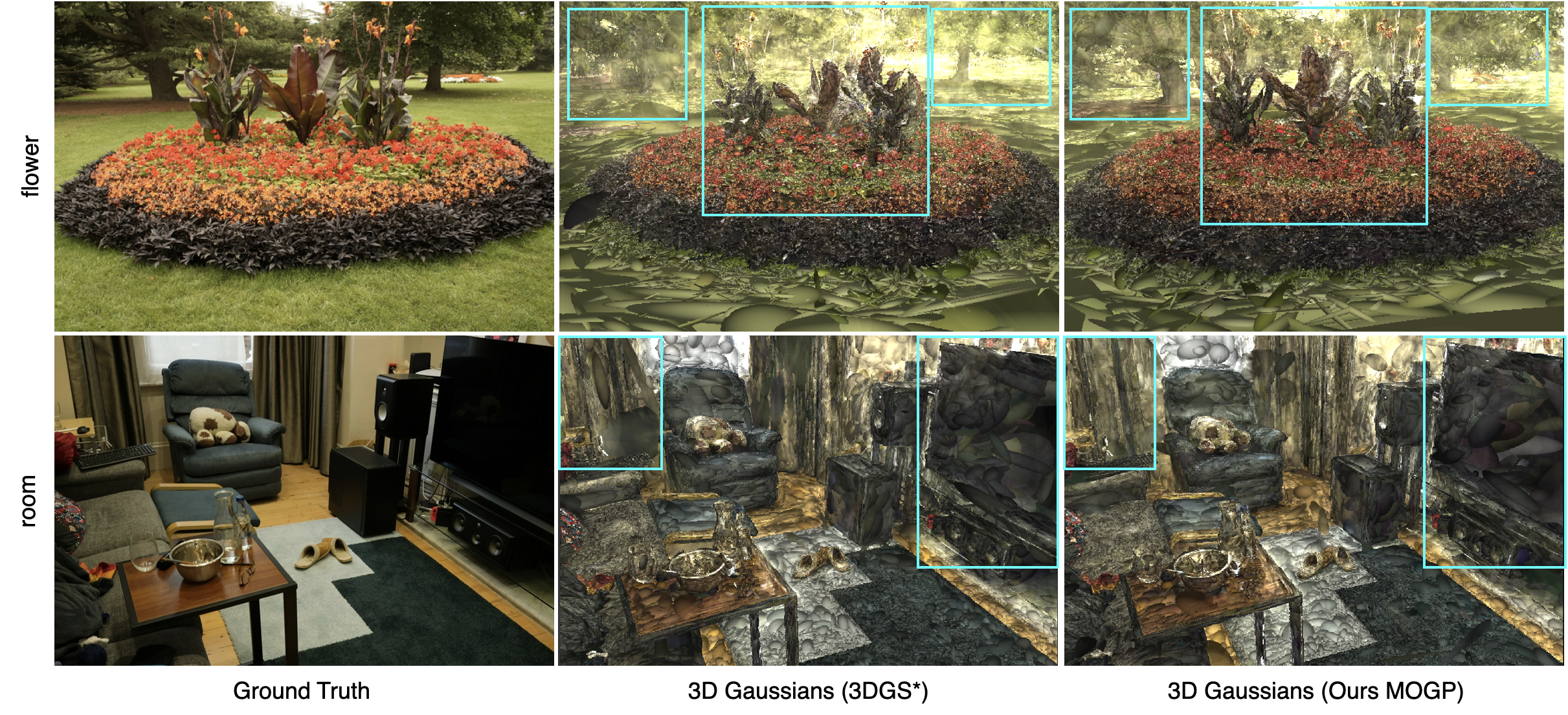}
    \caption{Qualitative comparison of 3D Gaussians during 3DGS training process. The left column shows the ground truth images. The middle column presents 3D Gaussians from 3DGS*, while the right column shows 3D Gaussians generated using ours MOGP.}
    \label{fig: comparison_3dgaussians}
\end{figure}
Further validating our densified point cloud, we present Figure~\ref{fig: comparison_3dgaussians}, which shows 3D Gaussians during training. Our approach leads to more structured and coherent reconstructions, especially in complex regions with fine textures and occlusions. In the flower scene, our 3D Gaussian representation successfully reconstructs branches, trunks, and leaves on the left and right sides, while the original 3D Gaussian sphere remains blurry and lacks structural detail. Additionally, our method accurately preserves the flower shape in the middle, improving overall reconstruction fidelity. In the room scene, our method provides a clearer representation of the curtain folds on the left, whereas the original method produces irregularly shaped Gaussian spheres and visible artefacts. The highlighted insets further demonstrate that our method reduces visual artefacts and better preserves geometric consistency.

\end{document}